\documentclass[letterpaper, 10 pt, conference]{ieeeconf}  % Comment this line out if you need a4paper

\IEEEoverridecommandlockouts                              % This command is only needed if 
\usepackage{amsmath} % assumes amsmath package installed
\usepackage{amssymb}  % assumes amsmath package installed
\usepackage[utf8]{inputenc}
\usepackage[keeplastbox]{flushend}
\usepackage{graphicx}
\usepackage{cite}
\usepackage{siunitx}
\usepackage[caption=false]{subfig}
\usepackage{afterpage}
\usepackage{booktabs}
\usepackage{siunitx}
\usepackage{multicol}
\usepackage[bookmarks=true]{hyperref}
\usepackage{mathdots} %for iddots

\title{Spherical Multi-Modal Place Recognition for Heterogeneous Sensor Systems
\author{Lukas Bernreiter, Lionel Ott, Juan Nieto, Roland Siegwart and Cesar Cadena} 
\thanks{This work was supported by the National Center of Competence in Research (NCCR) Robotics through the Swiss National Science Foundation.}
\thanks{All authors are with the Autonomous Systems Lab, ETH Zurich, Zurich 8092, Switzerland, {\tt \small \{berlukas, lioott, nietoj, rsiegwart, cesarc\}@ethz.ch.}}%
\thanks{\textcopyright 2021 IEEE. Personal use of this material is permitted. Permission from IEEE must be obtained for all other uses, in any current or future media, including reprinting/republishing this material for advertising or promotional purposes, creating new collective works, for resale or redistribution to servers or lists, or reuse of any copyrighted component of this work in other works
}%
}% <-this % stops a space

\usepackage[utf8]{inputenc}
\usepackage[OT1]{fontenc}

\usepackage{fancyhdr}

\usepackage{textcomp}\usepackage{gensymb}

\usepackage{multicol}

\usepackage{subfig}
\usepackage{graphicx}

\usepackage{upgreek}

\usepackage{units}

\usepackage{pdfpages}
\includepdfset{pages={-}, frame=true, pagecommand={\thispagestyle{fancy}}}

\usepackage[nolist,nohyperlinks]{acronym}

\newcommand{\norm}[1]{\left\lVert#1\right\rVert}

\renewcommand{\vec}[1]{\ensuremath{{\boldsymbol{#1}}}}

\begin{acronym}
\acro{SLAM}{Simultaneous Localization And Mapping}
\acro{TSDF}{Truncated Signed Distance Field}
\acro{ICP}{Iterative Closest Point}
\acro{RFS}{Random Finite Sets}
\acro{CNN}{Convolutional Neural Network}
\acro{CDF}{Cumulative Distribution Function }
\end{acronym}

\begin{document}

\maketitle

% 0 - Abstract
\begin{abstract}
In this paper, we propose a robust end-to-end multi-modal pipeline for place recognition where the sensor systems can differ from the map building to the query.
Our approach operates directly on images and LiDAR scans without requiring any local feature extraction modules.
By projecting the sensor data onto the unit sphere, we learn a multi-modal descriptor of partially overlapping scenes using a spherical convolutional neural network. 
The employed spherical projection model enables the support of arbitrary LiDAR and camera systems readily without losing information.
Loop closure candidates are found using a nearest-neighbor lookup in the embedding space.
We tackle the problem of correctly identifying the closest place by correlating the candidates' power spectra, obtaining a confidence value per prospect. 
Our estimate for the correct place corresponds then to the candidate with the highest confidence. 
We evaluate our proposal w.r.t. state-of-the-art approaches in place recognition using real-world data acquired using different sensors.
Our approach can achieve a recall that is up to $10\,\%$ and $5\,\%$ higher than for a LiDAR- and vision-based system, respectively, when the sensor setup differs between model training and deployment. 
Additionally, our place selection can correctly identify up to $95\,\%$ matches from the candidate set.
\end{abstract}

%\begin{IEEEkeywords}
% Localization, Mapping, Multi-Modal Perception
%\end{IEEEkeywords}
%
%\IEEEpeerreviewmaketitle
%
\section{Introduction}
Place recognition in mobile robotics with heterogeneous sensory systems are particularly challenging for current vision and laser-based place recognition systems.
Many of the existing approaches are often tailored to a specific LiDAR or camera and often do not generalize well to another system with different resolutions, variations in densities, or camera lenses. 
Furthermore, the problem becomes significantly more challenging for learning-based methods when the employed sensors are switched between training and testing or building and querying a map.

Such scenarios are especially crucial for autonomous vehicles as it enables building a generic model using high-performance sensors and test on manufacturer-specific, low-cost sensory systems.
Consequently, allowing manufacturers to freely choose the employed sensors and decrease their production costs using lower fidelity sensory systems.
Additionally, heterogeneous models would remain applicable even when the hardware is upgraded and would not be specialized for a specific year or vehicle generation. 
However, it is common for state-of-the-art place recognition systems to simplify the problem by assuming the same sensory system for training and testing as it is further, in many cases, inherently impossible to change sensor configurations without degradations once a model is learned.
\begin{figure}[!t]
  \centering
   \includegraphics[width=0.5\textwidth, trim={0.0cm, 0cm, 0.0cm, 0cm}, clip]{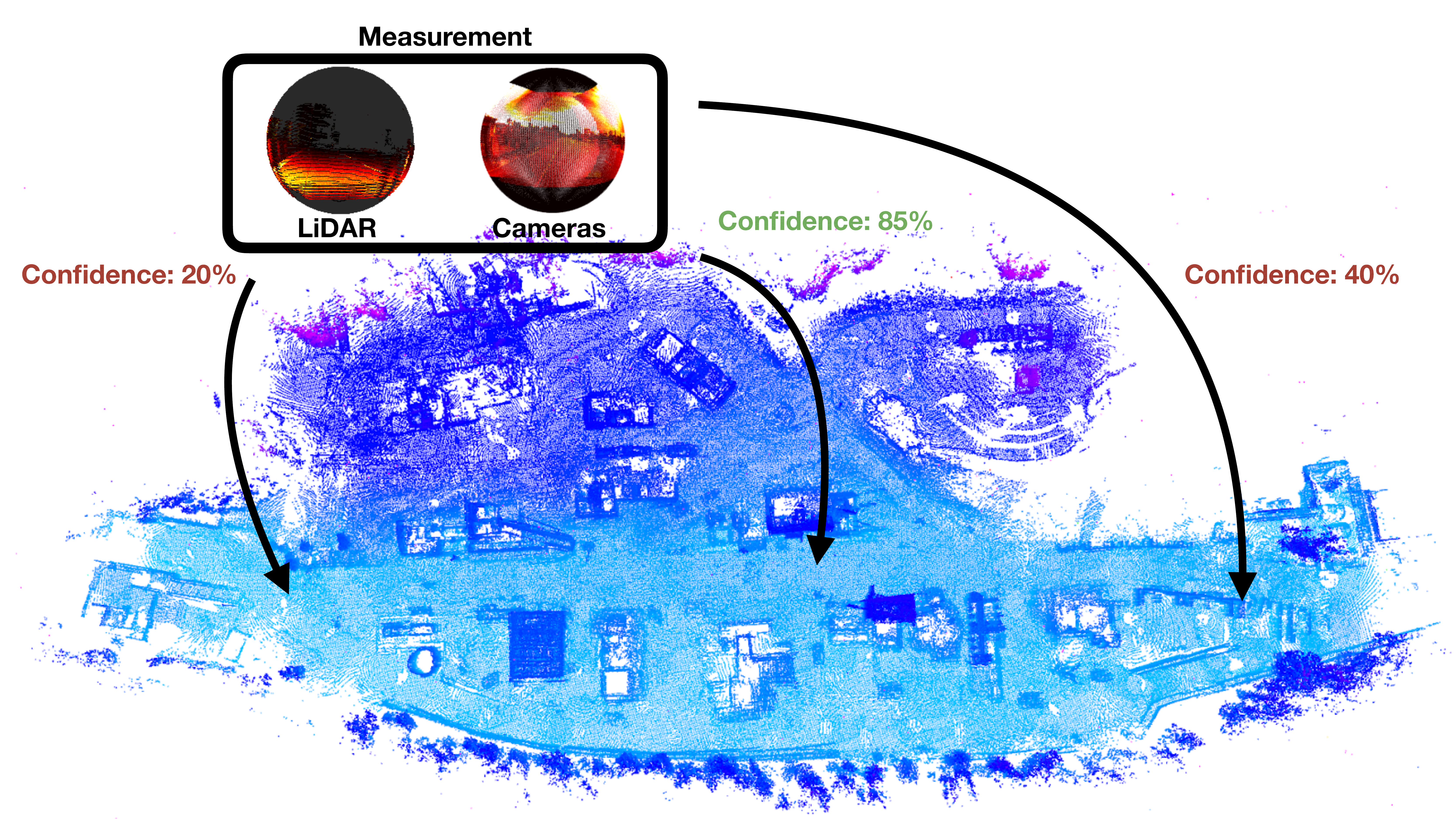}
   \vspace{-8mm}
   \caption{We propose a place recognition pipeline that combines, but is not limited to, the output of a LiDAR scan and multiple cameras. Furthermore, the outcome of our pipeline yields a confidence value for the place matching.}
   \label{pics:introduction:teaser}
   \vspace{-6mm}
\end{figure}
Moreover, a robust and reliable place recognition is often not achieved through a single sensory system, e.g., GNSS-based localization is insufficiently accurate in various scenarios due to multipath effects.
Image-based place recognition systems are prone to degradation when the data contains viewpoint and illumination changes and, further, commonly do not support multi-camera systems very well.
Similarly, several LiDAR-based place recognition systems require computationally expensive preprocessing steps, e.g., ground and noise removal, and typically suffer from degraded performance with rotated scans from the same scene.
Despite these challenges, the current state-of-the-art either addresses the vision- or laser-based place recognition problem and do not take advantage of their complementary nature.
Visual sensors provide descriptive appearance information about the environment, whereas LiDAR sensors are useful to measure the range accurately and, when combined, can increase the robustness of localization or place recognition systems.
Regardless of the employed sensors, most place recognition algorithms can only reliably retrieve the correct place when considering a high number of potential candidates.
Thus, making outlier rejection algorithms or additional filter steps a critical requirement.
%
%Another possible avenue to tackle the heterogeneous global place recognition problem is to abstract the raw measurements to semantic information. 
%Generally, semantics mitigate the place recognition problem by being mostly invariant to viewpoint and employed sensor system.
%Thus, has the possibility to significantly increase a system's robustness.
%
While most of the currently proposed solutions rely on a single modality per network, our approach combines multiple modalities in a single network and forward pass.
The employed spherical projection model seamlessly allows supporting high field-of-view systems without introducing distortions.
In more detail, we use a spherical \ac{CNN}~\cite{Cohen2018a, Esteves2018, Perraudin2019} to learn an embedding optimized for place recognition taking the projected sensor data as input (cf. Figure~\ref{pics:introduction:teaser}).
We retrieve loop closure candidates using a nearest neighbor search in the embedding space.
Furthermore, we estimate each candidate's power spectrum and perform a correlation in the spherical harmonic domain to find the best match between all retrieved candidates. 
In summary, our contributions are
\begin{itemize}
    \item An end-to-end pipeline for place recognition using a spherical projection of cameras and LiDARs.
    \item A probabilistic voting framework for finding the correct match given a set of  potential candidates.
    %\item A detailed evaluation of several heterogeneous place recognition scenarios.
\end{itemize}
We conduct a detailed evaluation of several heterogeneous place recognition scenarios and sensor setups.
%In the remaining part of this paper, we will first review the related work in Section~\ref{sec:related_work}. 
%Then, introduce our multi-modal place recognition pipeline in Section~\ref{sec:method}. 
%In Section~\ref{sec:experiments}, we will validate the performance of our proposed place recognition approach.
%Finally, we conclude our work in Section~\ref{sec:summary}.
%
\section{Related Work}
\label{sec:related_work}
This section reviews the current state-of-the-art vision- and LiDAR-based localization and place recognition systems related to our pipeline, where we begin with individual solutions and then conclude with multi-modal approaches.
\subsection{Visual-based Approaches}
Many traditional visual place recognition systems follow the traditional bag-of-words paradigm, where aggregated local descriptors (e.g. SIFT and SURF) represent an image or place~\cite{Philbin2007, Galvez-Lopez2012}.
However, learning-based solutions have significantly boosted the field~\cite{Zheng2018, Arandjelovic2018} in recent years.
S\"{u}nderhauf et al.~\cite{Sunderhauf2015} and Chen et al.~\cite{Chen2017a} describe multiple spatial regions using \ac{CNN} features for place recognition.
The work of Hausler et al.~\cite{Hausler2019} proposes a novel fusion of image processing methods for increased robustness and performance.
%Radenovic~\cite{Radenovic2017} proposes a novel feature pooling approach for better performance but requires high-resolution images. 
%
%
%Several of these methods deal with the description of keypoints and often can only reliable localize a mobile robot given perceptually similar conditions and sensors.
%Many robotic scenarios inherently can not fulfill these requirements, and thus, a significant amount of work emphasizes on avoiding localization degradation when the data contains seasonal and viewpoint discrepancies~\cite{Dusmanu2019, Revaud2019, Luo2020}.
%The work of Revaud et al.~\cite{Revaud2019} and Luo et al.~\cite{Luo2020} aim to increase the matching performance for scenes with viewpoint and illumination differences by jointly learning a keypoint detector and descriptor.
%Noh et al.~\cite{Noh2017} presented an approach with an attention-based selection of local features optimized for large scale place recognition.
%In contrast to keypoint-based methods, end-to-end learning approaches neither focus on keypoint extraction nor description but rather learn a representation of the whole image optimized for the place recognition task.
PoseNet~\cite{Kendall2015} is an end-to-end learning system for global localization, whereas NetVLAD~\cite{Arandjelovic2018} learns end-to-end a global descriptor for place recognition using a VLAD layer.

Furthermore, since many robotic scenarios inherently contain different appearances between map and query images, a significant amount of work tackles seasonal and viewpoint discrepancies to avoid degradation~\cite{Torii2018,Gordo2017}.

Our proposed approach also learns a representation in an end-to-end manner but does not solely rely on visual images but also takes LiDAR scans as an additional input.

\subsection{LiDAR-based Approaches}
Although visual localization approaches are more mature than their LiDAR-based counterparts, the latter can outperform visual systems due to their illumination-invariance and 360-degree field-of-view. 
Early advances in LiDAR-based localization and place recognition explore point histograms over the whole cloud~\cite{Rusu2010, Rohling2015}.
However, histograms are inefficient under partially overlapping pointclouds, where they differ. 

Here as well, learning-based approaches have gained tremendous success in recent years. 
Dube et al.~\cite{Dube2020} propose that with the accumulation of multiple LiDAR scans to individual segments it can learn a more distinctive description of the environment.
The use of attention networks for place recognition was proposed by Zhang et al.~\cite{Zhang2019} to reweight local feature points.
The work of Du et al.~\cite{Du2020} proposes a relocalization pipeline to extract local descriptors, a score map, and a global descriptor in a single forward pass.

Several other approaches utilize a projection to the 2D Euclidean domain and consequently rely on 2D CNNs~\cite{Yin2018b,Chen2020, Schaupp2019a}.
LocNet~\cite{Yin2018b} learns a rotation-invariant global descriptor of pointclouds using histograms per scan line.
Uy et al.~\cite{Uy2018} propose a NetVLAD layer for description on top of PointNet~\cite{Qi2017a}. 
Recently, Chen et al.~\cite{Chen2020} propose an end-to-end learning approach, which additionally evaluates the matching based on the overlap. 
Some other global LiDAR-based localization approaches additionally utilize the returned signal's strength (intensities) and have shown improvements to the localization quality~\cite{Chen2020,Guo2019, Cop2018a}.  

While LiDAR-based approaches typically perform well in confined environments, their performance often suffers in wide open areas, due to the limited amount of returned beams.
Consequently, the combination of multiple modalities on a fundamental level is paramount to design a robust and reliable place recognition system for arbitrary environments.
Thus, our approach aims at combining the visual and LiDAR-based place recognition into a single pipeline.

\subsection{Multi- and Cross-modal Approaches}
Recently, there is great interest in robotics to perform multi- or cross-modal place recognition and localization.
In the work of Ratz et al.~\cite{Ratz2020}, a multi-modal descriptor was learned by fusing the embedding of a NetVLAD and of LiDAR segments~\cite{Dube2020} using fully connected layers.
Xie. et al.~\cite{Xie2020} presented an end-to-end approach with an image and pointcloud fusion to learn a multi-modal descriptor.

The work of Cattaneo et al.~\cite{Cattaneo2019} and  Caselitz et al.~\cite{Caselitz2016} deal with the task of localizing visual data on LiDAR maps.
Feng et al.~\cite{Feng2019} deals with the extraction of descriptors of images and LiDAR patches in a metric learning approach.

Inspired by these findings, we propose an end-to-end place recognition pipeline that combines vision and LiDAR sensors.
Both modalities are projected onto the unit sphere and are given as an input to a spherical \ac{CNN}~\cite{Cohen2018a, Esteves2018, Perraudin2019}. 
%As the work of Cohen et al.~\cite{Cohen2018a} and Esteves et al.~\cite{Esteves2018} have shown, spherical \ac{CNN}s work very well for several classification tasks, especially when dealing with rotated data.
As a result, our network learns a global descriptor directly from the input modalities and does not require any local feature extraction modules.
\begin{figure*}[!t]
  \vspace{0.2cm}
  \centering
   \includegraphics[width=0.95\textwidth, trim={0.0cm, 0cm, 0.0cm, 0cm}, clip]{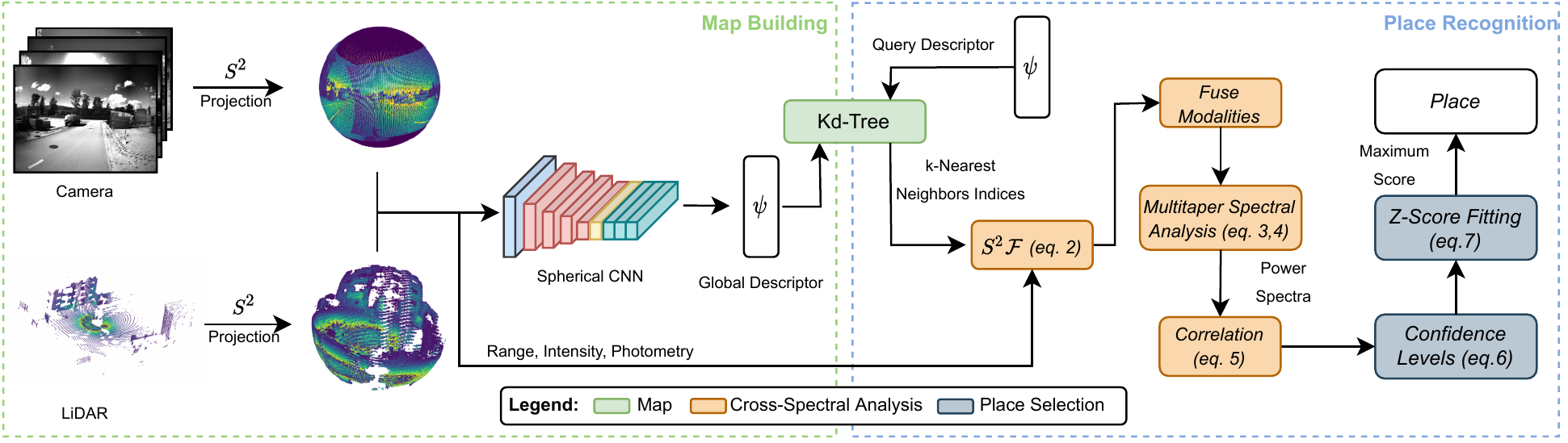}
   \vspace{-3mm}
   \caption{Overview of the proposed place recognition framework. Spherical projections of multiple images and a LiDAR scan serve as input for a spherical \ac{CNN}. The network learns to differentiate places through a global representation of the input modalities. The nearest neighbors are transformed using a spherical Fourier transform ($S^2\mathcal{F}$), fused, and then correlated. The correct place corresponds to the neighbor with the highest confidence in the correlation to the query. The related equations referenced by the components are discussed in Section~\ref{sec:method}.}
   \label{pics:method:overview}
   \vspace{-5mm}
\end{figure*}
\section{Method}
\label{sec:method}
This section describes the core components of our place recognition framework (cf. Figure~\ref{pics:method:overview}). First, we discuss the spherical projection model and how the modalities are combined. Second, we explain the spherical \ac{CNN} architecture that takes the spherical signals as input and learns a global descriptor. These two parts form the map building stage of our proposed methodology.
Finally, we describe our spectral analysis-based candidate selection and matching process.
\subsection{Data Representation}\label{sec:method:data_rep}
Overall, our approach utilizes multiple camera images and LiDAR scans to learn a global multi-modal descriptor of all modalities. 
Although we only use two modalities (camera and LiDAR) in this work, our approach is not limited to these and can readily be extended to include more.
We start by combining all modalities into one single joint representation to create an input feature vector for our network.

Since our pipeline's core component is a spherical \ac{CNN} that takes arbitrary square-integrable functions on the two-dimensional hypersphere $S^2$ as input, we project all modalities onto the sphere.
%We use the standard parametrization of the sphere, i.e. $\vec{\omega}(\phi,\theta)=[\cos\phi\sin\theta, \sin\phi\sin\theta,\cos\theta]^\top$ for $\vec{\omega}\in S^2$, where $\phi\in[0,2\pi]$ is the azimuth angle and $\theta\in[0,\pi]$ the inclination.
In addition, the spherical \ac{CNN} requires discrete equiangular samples and generally the input has to comply with Discroll and Healy's (DH) sampling theorem~\cite{Driscoll1994}.
Hence, we uniformly sample the projected modalities using an equiangular DH grid by performing a k-nearest neighbor lookup of the sampling points. 

Generally, the DH grid is defined as a $2\tilde{B}\times2\tilde{B}$ sampling grid, where $\tilde{B}$ is the spherical bandwidth which controls how dense the sphere is sampled and the discretization of the spherical spectrum. 
Using $\tilde{B}=100$, we define a DH grid in a common base frame enabling us to combine all the individual modalities.
%   In this work, the IMU acts as a common base frame between all employed sensors.
%Our approach essentially requires that all input sensors have a known extrinsic calibration to the base frame. 
The constructed equiangular sampling grid in the base frame is then projected into each sensor's local frame to sample the modality using each projected point that falls within the sensor's field-of-view.
For camera images, this corresponds to the photometric pixel value, and for LiDAR scans, we sample the range and intensity values. 
Multiple sensors per modality (e.g. multiple cameras) use the same sampling grid such that overlapping regions can be averaged when sampled. 
Regions that do not have a corresponding measurement are set to zero.
%We implemented the sampling using a nearest neighbor search for both, pointclouds and images. 
Finally, in the base frame, we combine all samples in our input feature vector $C\in\mathbb{R}^{3\times200\times200}$ comprising photometry, range, and beam intensity and forward it to the spherical \ac{CNN}.
\subsection{Spherical Convolutional Neural Network Architecture}\label{sec:method:s2cnn}
\begin{figure}[!t]
  \vspace{-2mm}
  \centering
   \includegraphics[width=0.5\textwidth, trim={0.0cm, 0.0cm, 0.0cm, 0.2cm}, clip]{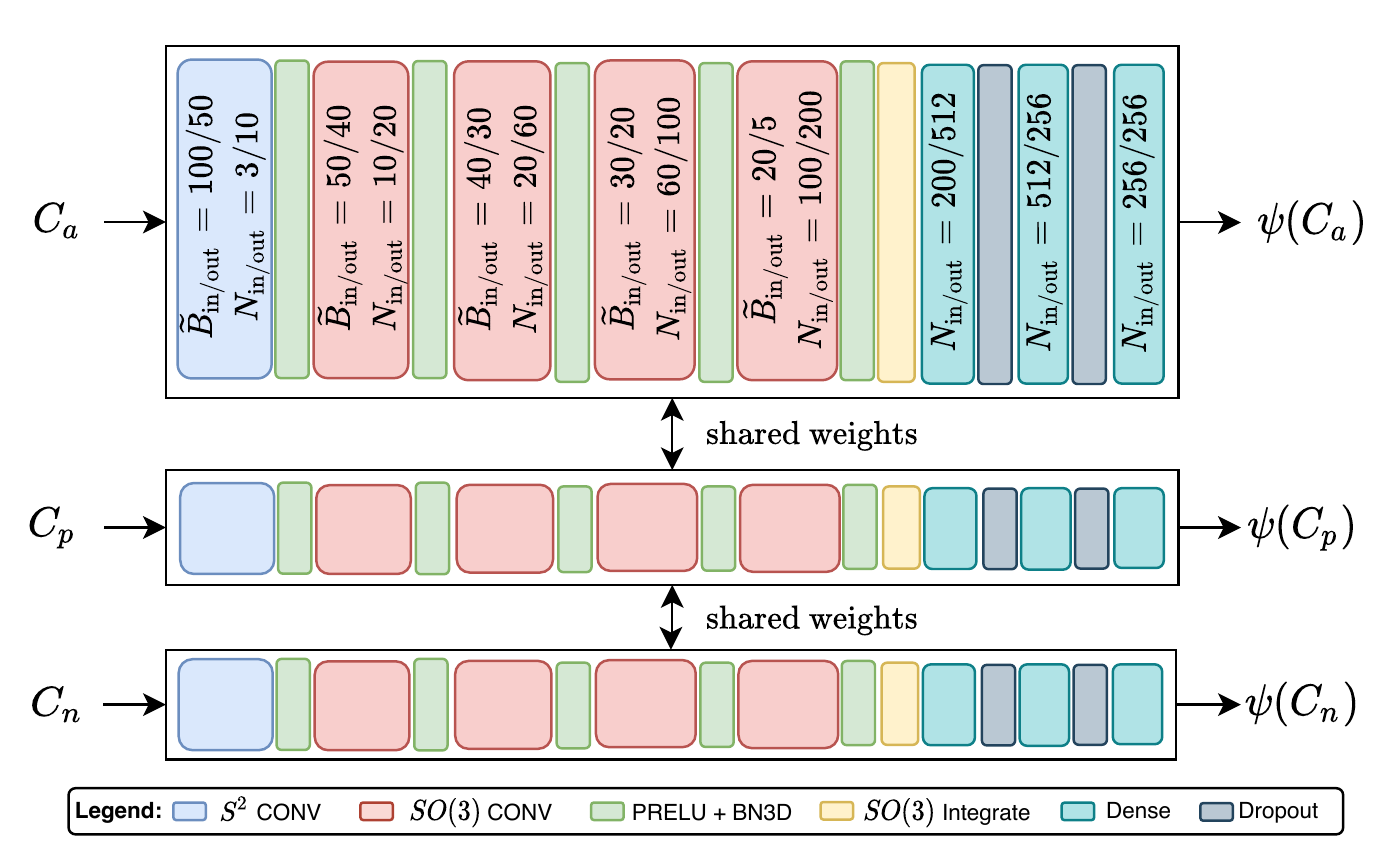}
   \vspace{-6mm}
   \caption{Detailed network architecture used to train our place recognition pipeline. Each training triplet $C_a$, $C_p$ and $C_n$ is passed through a duplicated network with shared weights.}
   \label{pics:methods:CNN_Structure}
   \vspace{-5mm}
\end{figure}
We utilize a spherical \ac{CNN}~\cite{Cohen2018a, Esteves2018, Perraudin2019} to learn a unique embedding of the multi-modal input data. 
As the work of Cohen et al.~\cite{Cohen2018a} and Esteves et al.~\cite{Esteves2018} have shown, spherical \ac{CNN}s work very well for several classification tasks, especially when dealing with rotated data.
Notably, for LiDAR-based systems, loop closure candidates can have arbitrary rotations between them, making spherical \ac{CNN}s a good fit.
Furthermore, several modern LiDAR systems leverage a high field-of-view resulting in high distortions for planar projection models. 
In contrast, spherical projections ideally model the nature of a rotating LiDAR and do not introduce any distortions into the projected pointcloud.

Our network architecture consists of $900k$ parameters, five convolutional and three fully-connected layers (cf. Figure~\ref{pics:methods:CNN_Structure}).
The first layer performs a convolution over $S^2$, whereas the remaining four convolutional layers act on $SO(3)$ to preserve the convolution's equivariance property~\cite{Cohen2017}. 
%We facilitate the spherical and $SO(3)$ Fourier transform~\cite{Kostelec2008} to perform each convolution efficiently.
Moreover, after each convolution, we apply a PReLu~\cite{He2015} activation function followed by a three-dimensional batch normalization (BN3D). 
After all convolutions, we integrate over $SO(3)$ and feed the result into our network's final three fully connected layers.
Between the fully connected layers, we employ dropout layers with a probability of $40\,\%$.
Finally, the outcome of the network will be a $256$-dimensional descriptor.

%As discussed in Section~\ref{sec:method:data_rep}, the input to our network will be the sampled spherical signals. 
%Each modality is treated as a separate input feature such that the network can learn how to select and combine the best of each. 

We approximate a function $\psi$ that maps our input $C$ to a unique embedding $\psi(C)$ optimized for place recognition.
Thereby, we employ a metric learning approach to learn a compact descriptor using a triplet network setup. 
Consequently, each training sample comprises an anchor $C_a$, a positive $C_p$ and a negative $C_n$ sample.
A positive sample represents a spatially close area to the anchor, whereas a negative sample is a place with no or minimal overlap w.r.t. the anchor. 
The loss is defined to pull positive matches closer and negative matches further away in the embedding space~\cite{Cheng2016}, i.e.

\begin{align}\label{eq:method:loss}
    \mathcal{L} = \frac{1}{N}\sum^N&\left[\norm{\psi(C_a)-\psi(C_p)}_2-\norm{\psi(C_a)-\psi(C_n)}_2+\tau_1\right]_+ \nonumber\\
    &+ \frac{1}{N}\sum^N\left[\norm{\psi(C_a)-\psi(C_p)}_2 - \tau_2\right]_+,
\end{align}
where $[z]_+=max(0,z)$, $N$ the number of triplet samples and $\tau_1=2,\tau_2=0.2$ denote the margins in the embedding space.
We employ a stochastic gradient descent and reduced the initial learning rate to $0.0046$ to ensure stable conversion.
Furthermore, we use a batch size of 13 samples for the optimization as this is our maximum GPU allowance. 

During the map building step, a KD-tree with an $L_2$ norm of the learned descriptors serves as map representation. 
The initial step for the place recognition is to perform a k-nearest neighbor search given a query descriptor. 
%We utilize at least six neighbors for the online lookup stage, as we found this to be most consistent and robust.
Next, we retrieve the corresponding equiangular feature vectors and perform a cross-spectral analysis between each of the nearest neighbors and the query to find the correct match.
\subsection{Spectral Analysis and Voting}\label{sec:method:spectral}
In this step, we seek to identify the correct place from a set of k-nearest neighbors and reject all outliers. 
Each neighbor is transformed to the spherical harmonics domain where we fuse the modalities, correlate the fused spectra with the query, and finally evaluate the correlation.  

In more detail, this process performs a cross-spectral analysis given the spherical harmonic coefficients of each neighbor in the spherical harmonic domain. 
Therefore, we first perform a spherical Fourier transform of each neighbor and each modality using the equiangular feature vectors. 

Generally, any arbitrary function $\tilde{f}\in L^2(S^2)$ can be expanded in the base of the spherical harmonics, i.e.
\begin{equation}\label{eq:pre:fourier_expansion}
    \tilde{f}(\vec{\omega}) = \sum_{l\ge0}\sum_{m\le{}l} \tilde{F}_{lm} Y_{lm}(\vec{\omega}),
\end{equation}
where $\tilde{F}$ is the spherical Fourier transformed signal and $Y_{lm}$ are the so called spherical harmonics of degree $l\in\mathbb{N}_0$, order $m\in[0,l]\in\mathbb{N}_0$ and form an orthonormal basis over $L^2(S^2)$.
For details on the spherical Fourier transform, we refer to the seminal work of Kostelec et al.~\cite{Kostelec2008}.

Before performing the spectral analysis, we create a fused spectrum by combining each of the transformed modalities.
Thereby, we compute the spectrum of each modality and select the modality's coefficient for the fused spectrum with the highest local power per degree~\cite{Falk1999a}.
The fused spectrum serves then as an input to the remaining part of the pipeline.

The correlation of two functions yields a measure of how strongly two functions are related and forms our basis for finding the correct match.
Therefore, at this point we estimate the power spectrum for each candidate and query.
Generally, the power spectrum $S_{\tilde{f}\tilde{f}}$ of a function $\tilde{f}$ is defined as the integral of $\tilde{f}$ squared over the spherical space.
Concretely, for a given degree $l$, the power spectrum of $\tilde{f}$ is given by
\begin{equation}\label{eq:method:power}
    S_{\tilde{f}\tilde{f}}(l)=\sum_{m=0}^l\tilde{F}_{lm}\cdot\tilde{F}_{lm}^{*}=\sum_{m=0}^l \operatorname{Re}(\tilde{F}_{lm})^2 + \operatorname{Im}(\tilde{F}_{lm})^2,
\end{equation}
where $\tilde{F}^*$ is the complex conjugate of $\tilde{F}$.
Similar, the cross power spectrum of two arbitrary functions $\tilde{f}$ and $\tilde{g}$ is defined by
\begin{equation}\label{eq:method:cross_power}
    S_{\tilde{f}\tilde{g}}(l) = \sum_{m=0}^l \tilde{F}_{lm} \cdot \tilde{G}_{lm}^{*}.
\end{equation}

Given the global power $S_{\tilde{f}\tilde{f}}$, $S_{\tilde{g}\tilde{g}}$ and the cross-power $S_{\tilde{f}\tilde{g}}$ spectra, we define the correlation $Q$ of two functions $\tilde{f}$ and $\tilde{g}$ for degree $l$ as
\begin{equation}\label{eq:methods:corr}
    Q(l) = \frac{S_{\tilde{f}\tilde{g}}(l)}{\sqrt{S_{\tilde{f}\tilde{f}}(l)\cdot{}S_{\tilde{g}\tilde{g}}(l)}}.
\end{equation}
The correlation in eq.~\ref{eq:methods:corr} forms the fundamental theory of a global spectral analysis in the spherical harmonic domain. 

Generally, a traditional, global Fourier cross-spectral analysis is often biased in terms of larger variances due to leakage. 
Thomson's pioneering work~\cite{Thomson1982} proposes an approach for alleviating biases by using multiple localized windows (tapers) to relate to the global spectrum. 
The main idea is to average a set of direct spectrum estimators using pairwise orthogonal tapers resulting in a less biased estimate of the power spectrum.
Generalizing the theory of the Cartesian multitaper analysis to the sphere~\cite{Wieczorek2007, Simons2006} enables our approach to examine the place candidates based on their cross-spectral energy. 
For a detailed derivation of the spherical tapers, we refer the interested reader to the work of Wieczorek et al.~\cite{Wieczorek2005, Wieczorek2007}.

Initially, we create a set of tapers $h_1,...,h_n$ to use during the complete candidate voting process. 
We first calculate the spectra $S_{\tilde{f}\tilde{f}}$, $S_{\tilde{g}\tilde{g}}$ and $S_{\tilde{f}\tilde{g}}$ using eq.~\ref{eq:method:power} and eq.~\ref{eq:method:cross_power}, respectively.
Next, for each taper $h_i$, we calculate the respective windowed version of the global power spectra and cross-spectrum~\cite{Wieczorek2005}.
%In various work~\cite{Wieczorek2005, Hivon2002} it was shown that the expectation of a windowed power spectrum relates to the true global power spectrum and can estimated with less bias by averaging multiple windowed version of it. 
Subsequently, using the windowed spectra, we compute the correlation $Q$ as in eq.~\ref{eq:methods:corr} for each $h_i$. 
The average over all correlations yields our measure on how related the query and the candidate are.

As a final step, we infer a confidence value based on the averaged correlation.
Generally, the candidate with the highest total confidence is selected as the final estimate for the robot's place. 
The probability that two functions correlate is expressed as bivariate normal distribution~\cite{Pauer2006} using the correlation coefficient $Q(l)$, i.e. 
\begin{align}\label{eq:methods:confidence}
    G_1(Q, 1) &= Q(1) \nonumber \\
    G_l(Q, l) &= Q_{l-1} + Q(l)(1-Q(l)^2)^{l-1}\prod_{i=1}^{l-1}\frac{2i-1}{2i}.
\end{align}
Analogous to a standard Fourier transform, most of the transformed signal's energy is concentrated in the lower bands.
Hence, we solely utilize the first 15 spherical harmonic degrees to evaluate a correlation coefficient in eq.~\ref{eq:methods:corr} and consequently, the confidence using eq.~\ref{eq:methods:confidence}. 
The resulting confidence values are converted to z-score values using the inverse of the \ac{CDF} of a normal distribution. 
In more detail, given a confidence value $g\in\left[0,1\right]$ we infer the z-score $s_g$ using
\begin{equation}
    s_g = \Phi^{-1}\left(\frac{1-(1-g)}{2}\right),
\end{equation}
where $\Phi$ is the \ac{CDF} of $\mathcal{N}(0,1)$.
The z-score down-weighs correlations with less than $50\,\%$ and up-weighs correlations with more than $50\,\%$ confidence.
We accumulate the z-scores for each degree $l$ and the maximum of the accumulated values represents the place with the highest confidence based on the correlation and consequently our best match.

Thus, the spectral analysis and place voting constitutes our proposed place recognition pipeline.
%If required, as a final step, the concrete localization can be computed using a pointcloud registration using the query and the best match.
%
\section{Experiments}
\label{sec:experiments}
This section evaluates our proposed spherical place recognition pipeline denoted as \textit{$S^2$Loc} using different sensory systems and environments. 
First, we discuss our used sensor setup and the data generation process. 
Then, the experimental evaluation of the learned descriptor and our proposed spectral place voting are presented. 

We compare our approach to two current state-of-the-art solutions: NetVLAD~\cite{Arandjelovic2018} for image-based place recognition and OverlapNet~\cite{Chen2020} for laser-based place recognition.
NetVLAD was set to treat each image independently for multi-camera datasets, and we configured OverlapNet to get the same information as our network, i.e., to use range and intensity information solely. 
\subsection{Sensor Setup and Data Generation}
Generally, for each test environment, we created a global multi-session map from multiple runs and jointly optimized it using constraints from visual landmarks, LiDAR scan-to-scan matches, and RTK GPS.
Our high-fidelity (HF) sensor setup comprises an Ouster OS-0 with 128 beams (131072 points per scan) and four $0.4\,\mathrm{MP}$ global shutter grayscale cameras (two forward, one to each side). 
The low-fidelity (LF) setup contains an Ouster OS-1 with 64 beams (65535 points per scan) and a single forward-facing $0.4\,\mathrm{MP}$ global shutter grayscale camera.

As discussed in Sec.~\ref{sec:method:s2cnn}, our triplet network approach requires three samples per training input, an anchor, a positive and a negative sample.
Positive candidates for training are extracted using a proximity search on the individual poses. 
Precisely, we extract positive intra- and inter-mission poses for each pose if their Euclidean distances are less than $5\,\mathrm{m}$. 
Similar, negative training candidates are extracted with a Euclidean distance of $6\text{-}20\,\mathrm{m}$.
Furthermore, we avoid clusters in the training and test data by constraining each sample to be at least $10\,\mathrm{cm}$ away from each other.
OverlapNet and NetVLAD were given the same training data as our network. 

Moreover, the data used for training and testing comprises an outdoor and indoor environment. 
Both environments were recorded with both sensor configurations using a handheld device~\cite{Tschopp2020}.
The recordings of the LF and HF setup are roughly one year apart for both datasets. 

\textbf{Outdoor environment.} We utilized a search and rescue testing facility (cf. Figure~\ref{pics:introduction:teaser}) that includes several multi-floor buildings, urban-like streets, and collapsed structures.
The training, test split was $8\,\mathrm{km}$/$2\,\mathrm{km}$ and $2\,\mathrm{km}$/$1\,\mathrm{km}$ for the HF and LF map, respectively.
Additionally, we ensured that the training and test data was not recorded on the same day.

\textbf{Indoor environment.} In total, the data covers $1\,\mathrm{km}$ of recordings in a building and is solely used for cross-modality tests, i.e., a HF map and LF queries. This data is never used for training.

\textbf{Evaluation metric.}
We use the \textit{recall@n} metric as the basis of our evaluation and comparison to the other approaches. 
Here, \textit{n} refers to the number of nearest neighbors (database candidates) retrieved during the lookup, and \textit{recall} refers to the correctly identified places w.r.t. the whole map.
For OverlapNet, we created a pointcloud map and retrieved the top-n candidates in terms of their highest overlap.
%A successful candidate match is found when the relative distance between the poses is less than $5\,\mathrm{m}$. 
%
\subsection{Descriptor Lookup}
This section investigates the accuracy and precision of our learned descriptor. 
Specifically, for a given descriptor map, we perform a nearest neighbor lookup and consider it as successful if one of the neighbors is within $5\,\mathrm{m}$ of the query sample.

\begin{figure}[!thb]
\vspace{-4.5mm}
\captionsetup[subfigure]{labelformat=empty}
\centering
    \subfloat {%
       \includegraphics[width=0.24\textwidth, trim={0.0cm, 0.1cm, 0.0cm, 0.1cm}, clip]{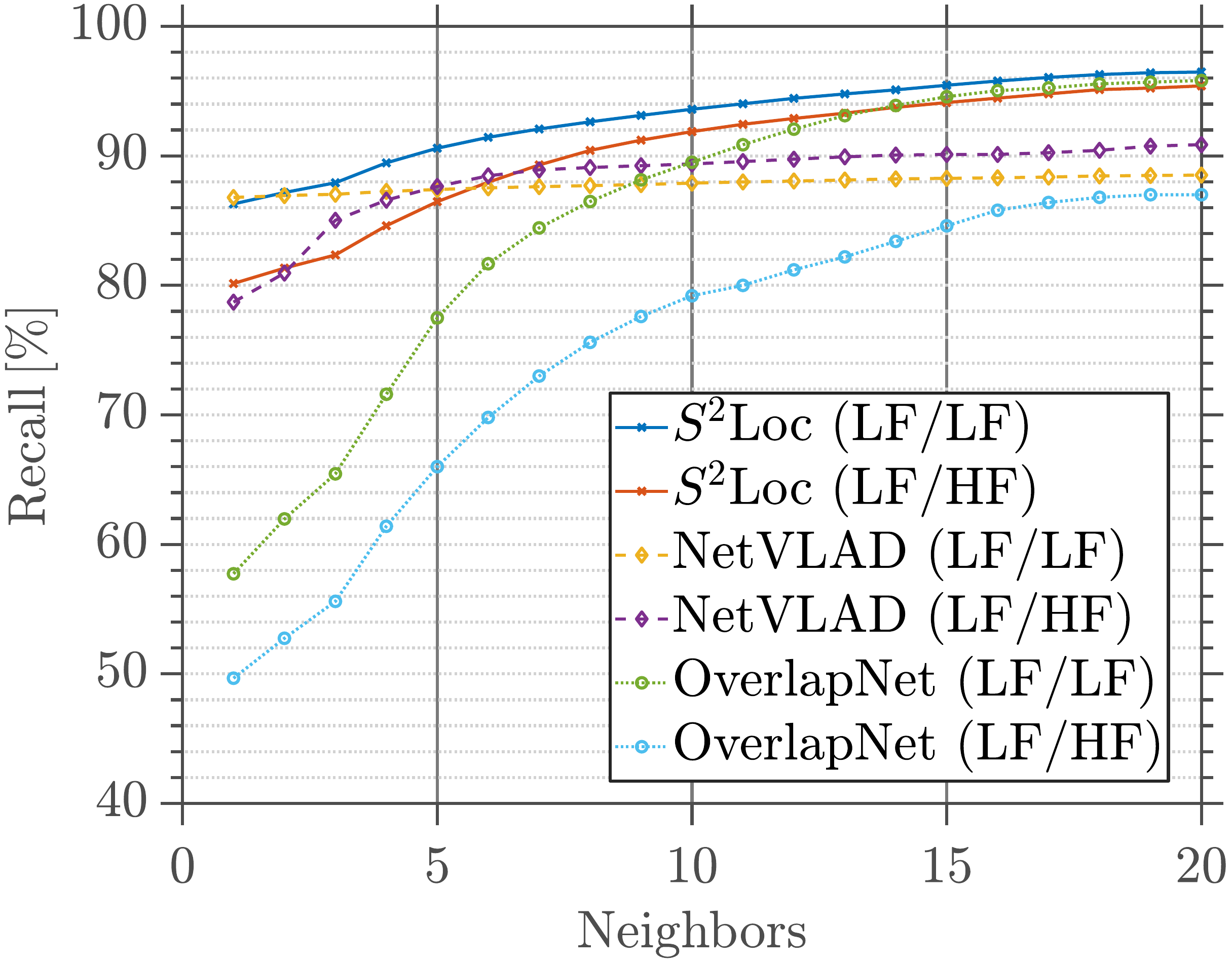}
    } 
    \subfloat {%
       \includegraphics[width=0.24\textwidth, trim={0.0cm, 0.1cm, 0.0cm, 0.1cm}, clip]{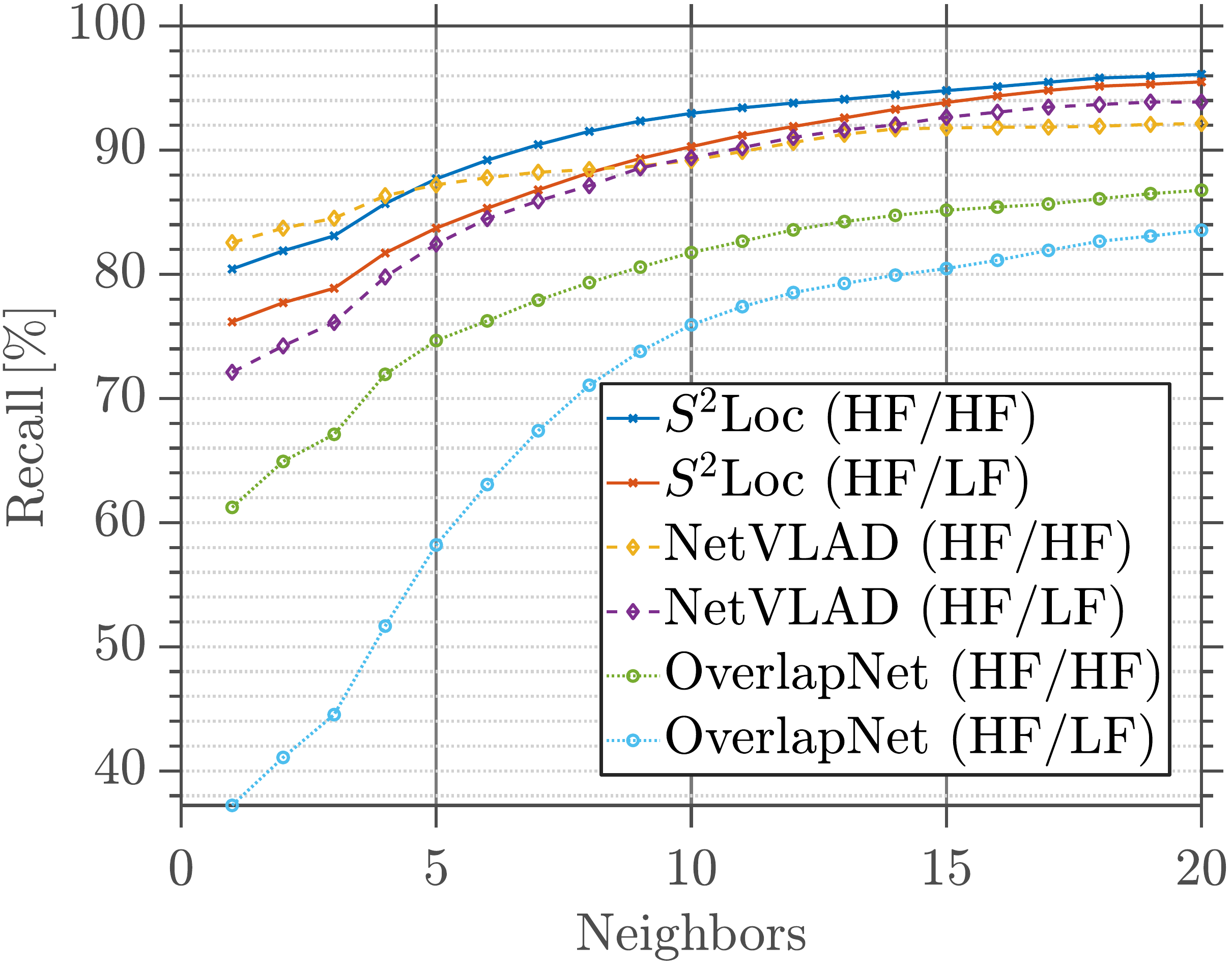}
    }
   \vspace{-3mm}
   \caption{Recall of the descriptor lookup in the outdoor environment. Each test set comprises 3000 places. The parentheses denote the used sensory system for training and testing, i.e. (train/test).  }
   \label{pics:exp:descriptor_lookup_recall_arche}
   \vspace{-2mm}
\end{figure}

\textbf{Descriptor matching.} 
This experiment aims to validate our descriptor's effectiveness when employing it on the same and different hardware from the outdoor environment (cf. Figure~\ref{pics:exp:descriptor_lookup_recall_arche}). 
In concrete, we compare different train and test configurations to validate the effectiveness of our multi-modal descriptor and show its versatile applicability. 
The test data is randomly sampled and contains arbitrary large rotational differences between map and query, making the LF data particularly hard for NetVLAD as there was only one camera used. 
As a result, NetVLAD's recall is competitive for less nearest neighbors but does not improve drastically for a higher number of retrievals. 
Furthermore, the data's diversity results in many samples exceeding our success threshold but are still within high overlap to several queries. 
Consequently, OverlapNet's recall results in a steady improvement with the number of retrieved neighbors. 
Our approach benefits from both modalities and, especially for more neighbors, generalizes well on the different sensory systems.
The visual data improves the retrieval when the LiDAR does not have reasonable good beam returns. 
Similarly, the LiDAR supports the retrieval when the visual counterpart suffers greatly from viewpoint and illumination changes.
Moreover, our approach gains additional efficiency from spherical CNN's rotational robustness, which we validate next along with the fact that our approach also works well with a single modality.

\begin{figure}[!htb]
\vspace{-2mm}
  \centering
   \includegraphics[width=0.44\textwidth, trim={0.0cm, 0cm, 0.0cm, 0cm}, clip]{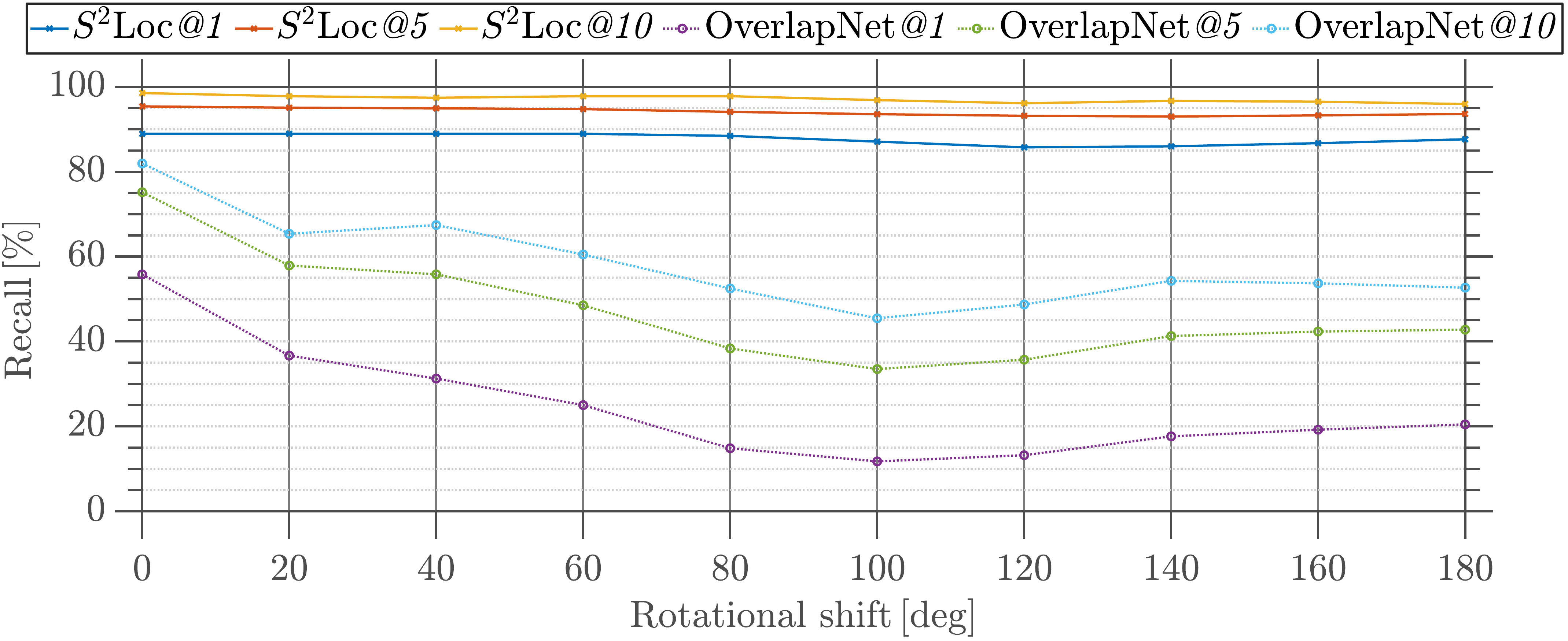}
   \vspace{-3mm}
   \caption{Evaluation of the descriptor lookup for the LiDAR-only case. The map was built with artificially rotated pointclouds and queries remained unchanged. Results are shown over a 1000 places with the \textit{recall@n} metric. The map was built and queried with the HF setup and both networks were trained on HF data.}
   \label{pics:exp:rotated_desc_lookup_recall}
   \vspace{-2mm}
\end{figure}

\textbf{Descriptor matching on rotated samples.} 
Next, we will investigate the descriptor matching when the data is corrupted with rotations.
Generally, in place recognition, LiDAR scans used to build and query a map can be arbitrary rotated to each other.
This experiment confirms that our approach is resilient against arbitrary rotations by corrupting the map with rotations around yaw.
%For testing, we rotated each map sample around yaw and left the query samples unrotated. 
Figure~\ref{pics:exp:rotated_desc_lookup_recall} illustrates the evaluation of rotational shifts from $0^{\circ}$ to $180^{\circ}$. 
These results indicate our spherical projection and the spherical \ac{CNN} generalize well on the rotated data and essentially account for almost no degradation in the recall.
In contrast, OverlapNet significantly decreases with increasing yaw perturbations since rotations around yaw result in shifts for planar projection models. 
%The results indicate that the learned multi-modal descriptor is suitable for place recognition.
Next, we test the place selection part of our proposed pipeline.
\subsection{Place Recognition}
\begin{figure}[!htb]
  \centering
   \vspace{-1mm}
   \includegraphics[width=0.44\textwidth, trim={0.0cm, 0cm, 0.0cm, 0cm}, clip]{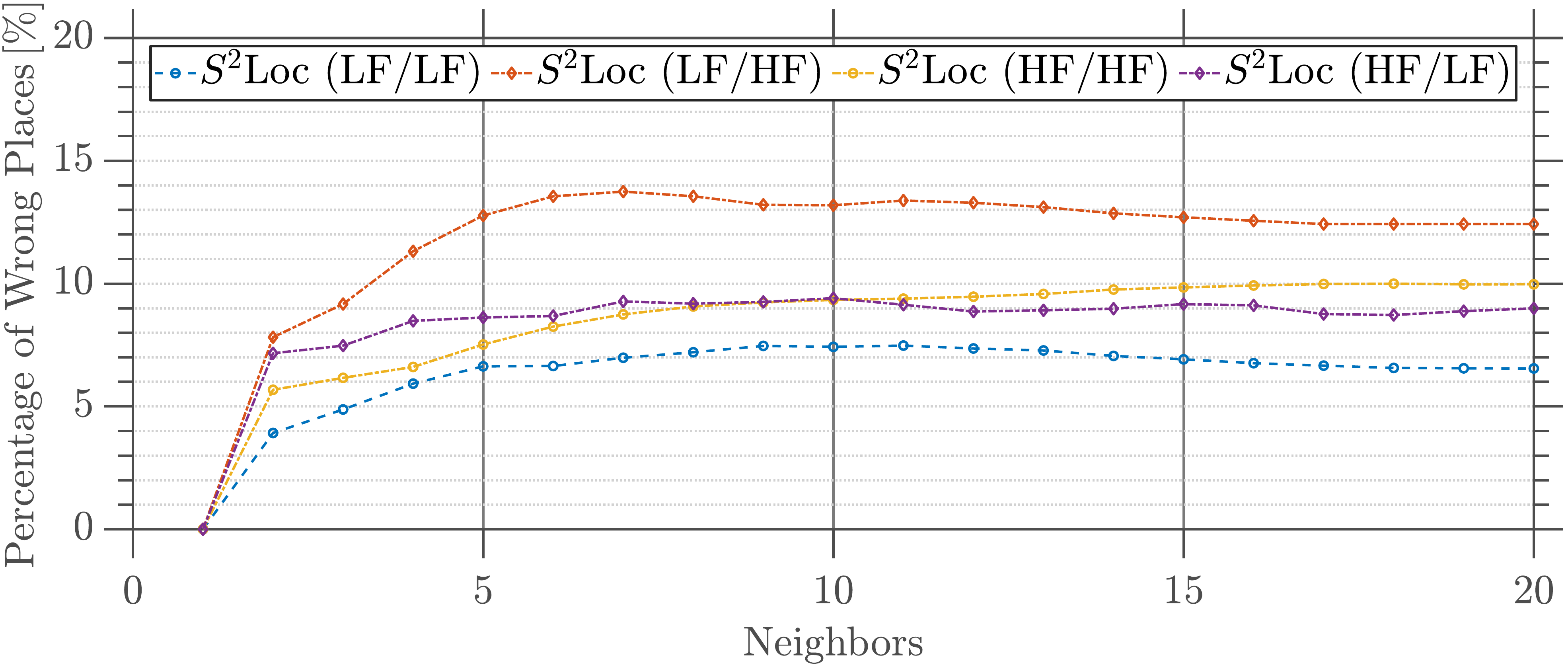}
   \vspace{-3mm}
   \caption{Percentage of wrongly selected places per database retrievals using our cross-spectral place selection described in Section~\ref{sec:method:spectral}. The results are shown for 2000 samples from the outdoor dataset.}
   \label{pics:exp:arche_taper_recall}
   \vspace{-2mm}
\end{figure}

This section solely evaluates our proposed cross-spectral place selection algorithm that takes place after the descriptor lookup.
Like the previous experiments, we only consider successful matches if the selected place from the z-score voting is within $5\,\mathrm{m}$ of the query location.
However, we limited the input to cases where at least one retrieved neighbor is a correct match to evaluate solely the place selection.
Figure~\ref{pics:exp:arche_taper_recall} shows the percentage of selected places greater than $5\,\mathrm{m}$ w.r.t. the number of retrieved candidates.
%The recall decreases for the first ten neighbors for all the tests but becomes steady for the remaining ten and does not reach below $80\%$. 
%For the same sensor and in the \textit{HR/LR} case, the probability is steady below $10\%$. 
Our approach considerably benefits from the descriptive information in the visual data to distinguish the correct from the wrong places.

\begin{figure}[!htb]
  \vspace{-1mm}
  \centering
   \includegraphics[width=0.47\textwidth, trim={0.0cm, 0cm, 0.0cm, 0cm}, clip]{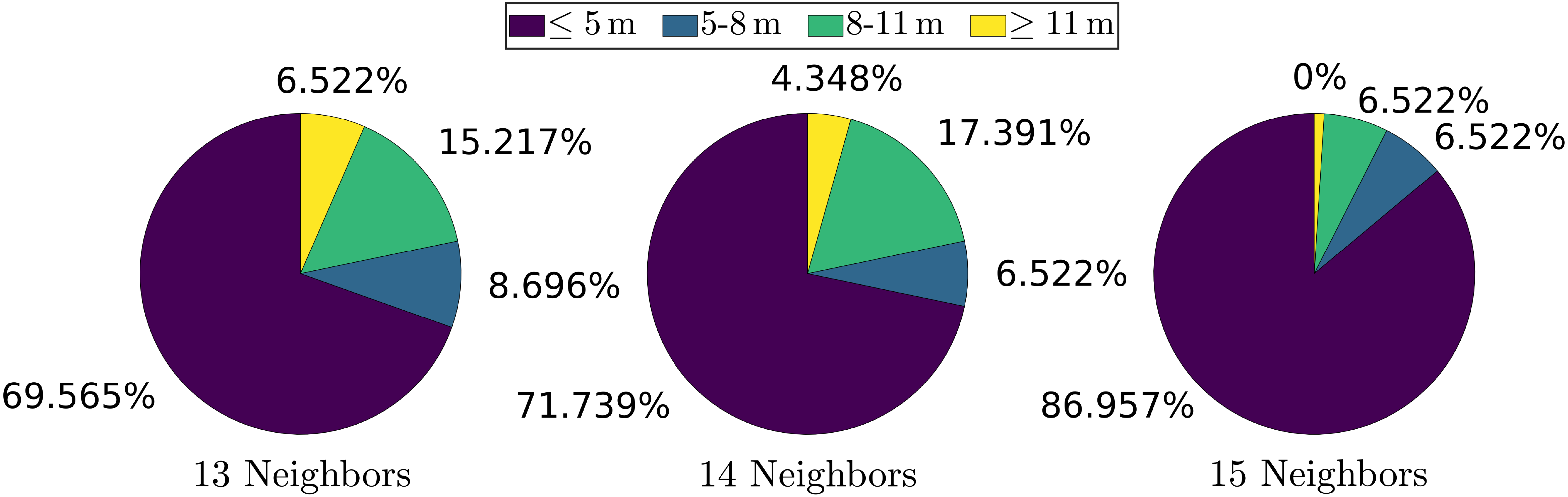}
   \vspace{-3mm}
   \caption{Distribution of the distances between the selected map places and their respective queries using 13, 14 and 15 neighbors. 
The map was built with HF and queried with LF sensors from the indoor dataset, whereas the model was trained on the HF outdoor dataset. }
   \label{pics:exp:cross_modality}
   \vspace{-3mm}
\end{figure}

Next, we evaluate our place selection using the cross-modal indoor environment where the sensors are switched between building (HF) and querying (LF) the map. 
Figure~\ref{pics:exp:cross_modality} illustrates the percentage of the selected places grouped by their distance to the query. 
The multi-modality and the correlation with less-biased spectrum estimates allow our approach to successfully distinguish close places as we are not computing correspondences but directly measure how similar the candidate with the query is.
Furthermore, with only 15 neighbors, no place further than $11\,\mathrm{m}$ away was selected. 
%Each model operates best on the sensor data that was also used during the training. 
\subsection{Performance Evaluation}
\begin{figure}[!htb]
  \vspace{-3mm}
  \centering
   \includegraphics[width=0.35\textwidth, trim={0.0cm, 0cm, 0.0cm, 0cm}, clip]{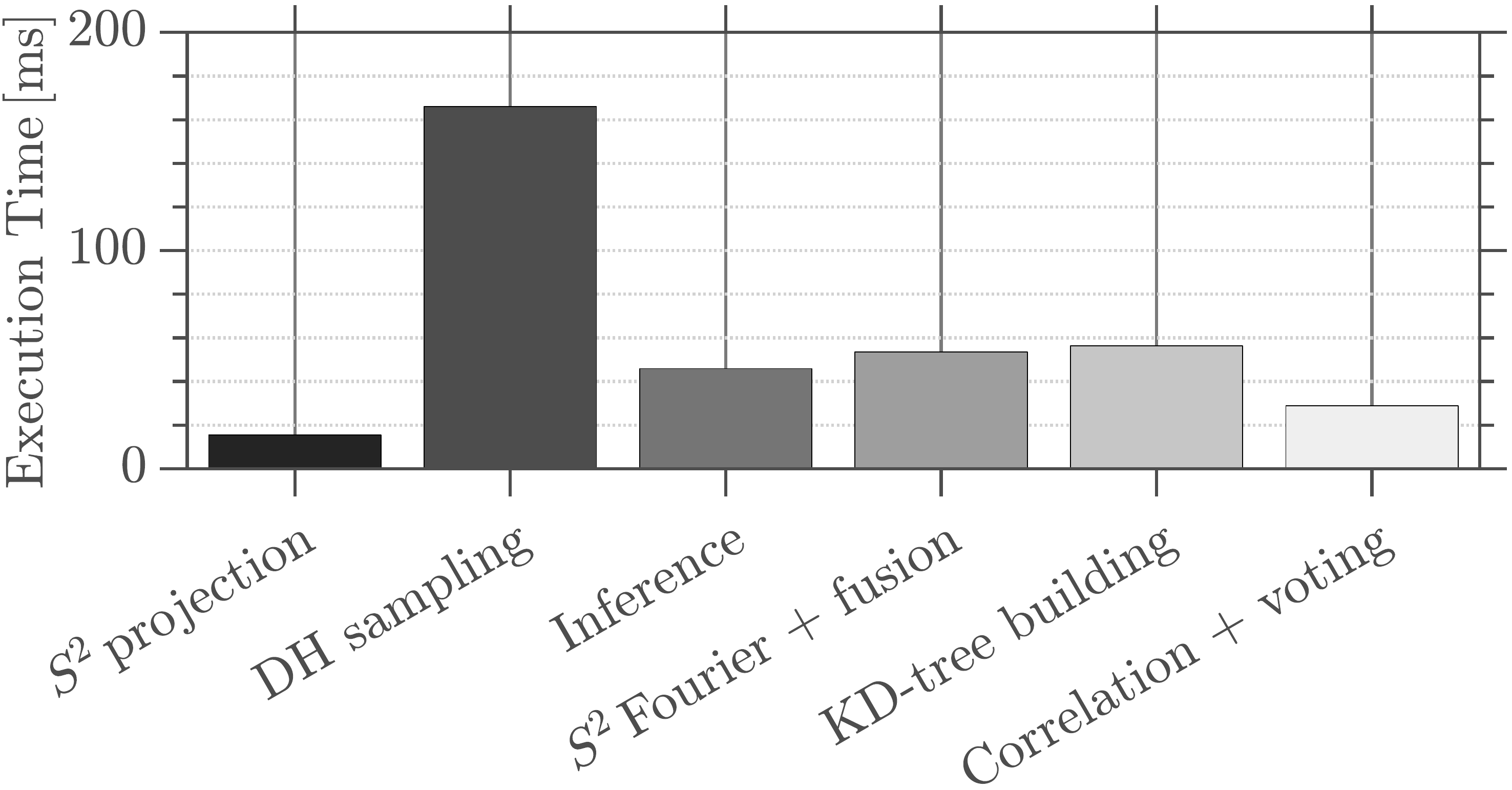}
   \vspace{-3mm}
   \caption{Execution time in $\mathrm{ms}$ partitioned per component. All values are averaged over a 1000 samples. }
   \label{pics:exp:performance}
   \vspace{-5mm}
\end{figure}

We run our proposed pipeline on an Intel Xeon E5-2640v3 and an NVIDIA Titan RTX. 
Figure~\ref{pics:exp:performance} shows the execution time per individual component. 
In total, a single sample takes \texttildelow{}$300\,\mathrm{ms}$ processing time for $\tilde{B}=100$ when considering $1.06\,\mathrm{ms}$ for the KD-tree lookup and disregarding the time needed to build the map.

\section{Conclusion and Future Work}
\label{sec:summary}
This paper presented a learning approach to multi-modal place recognition using vision and LiDAR sensors.
Our approach operates end-to-end by projecting each modality onto the hypersphere and using a spherical \ac{CNN}. 

We showed that our multi-modal descriptor improve the state-of-the-art in place recognition, and more important, it generalizes to different sensor systems, in terms of training and deployment, and map building and querying.
Additionally, our method benefits from spectral analysis to efficiently distinguish the correct place from the retrieved database candidates.

%As further research, we would like to explore semantic objects as a unique and descriptive representation of places. 
%Moreover, the characteristics of all currently used channels are inherently distinctive, and therefore, we intend to investigate the heterogeneous place recognition and spectral analysis more thoroughly and account for it.
%
We will continue our research in two directions. First, the integration of semantic information (e.g. computed from the image) as an extra modality to improve in the place recognition task. Second, we will investigate a spherical decoder to exploit our multi-modal spherical parametrization for the actual task of semantic segmentation under heterogeneous sensor coverages and sensor setups.
\addtolength{\textheight}{-1cm} 
%\section*{Appendix}
%\section*{Acknowledgment}
%This work was partially supported by the German Research Foundation (DFG) under grant HA 3789/13-1.
%
\bibliographystyle{IEEEtran}
\bibliography{bib/references.bib}

\end{document}